%% file: mlsp_template_camera_ready.tex
\documentclass{article}
\usepackage{amsmath,graphicx,bm,stmaryrd,amsfonts,amsmath,algorithm,color,caption,subcaption,authblk}

\usepackage[small,compact]{titlesec}

\twocolumn
\def\ninept{\def\baselinestretch{.95}\let\normalsize\small\normalsize}

\def\@maketitle{\newpage
 \null
 \vskip 3em \begin{center}
 {\large \bf \@title \par} \vskip 1.5em {\large \lineskip .5em
\begin{tabular}[t]{c}\@name \\ \@address
 \end{tabular}\par} \end{center}
 \par
 \vskip 1.5em}

\usepackage[noend]{algpseudocode}
\DeclareMathOperator*{\argmin}{arg\,min}
\input{pre-packages.tex}
\input{pre-declarations.tex}
\input{pre-definitions.tex}
\let\oldthebibliography\thebibliography
\renewcommand\thebibliography[1]{%
  \oldthebibliography{#1}%
  \setlength{\itemsep}{0pt plus 0.3ex}%
}

\usepackage{mathtools}
\mathtoolsset{showonlyrefs}

\title{A Time-aware tensor decomposition for tracking evolving patterns}

\author[1,2]{Christos Chatzis}
\author[3]{Max Pfeffer}
\author[2]{Pedro Lind}
\author[1]{Evrim Acar}
\affil[1]{Simula Metropolitan Center for Digital Engineering, Oslo, Norway}
\affil[2]{Faculty of Technology, Art and Design, OsloMet, Oslo, Norway}
\affil[3]{Institute for Numerical and Applied Mathematics, Georg-August-Universität Göttingen, Göttingen, Germany}
\date{}
\begin{document}

\maketitle

\begin{abstract}
Time-evolving data sets can often be arranged as a higher-order tensor with one of the modes being the time mode. While tensor factorizations have been successfully used to capture the underlying patterns in such higher-order data sets, the temporal aspect is often ignored, allowing for the reordering of time points. In recent studies, temporal regularizers are incorporated in the time mode to tackle this issue. Nevertheless, existing approaches still do not allow underlying patterns to change in time (e.g., spatial changes in the brain, contextual changes in topics). In this paper, we propose temporal PARAFAC2 (tPARAFAC2): a PARAFAC2-based tensor factorization method with temporal regularization to extract gradually evolving patterns from temporal data. Through extensive experiments on synthetic data, we demonstrate that tPARAFAC2 can capture the underlying evolving patterns accurately performing better than PARAFAC2 and coupled matrix factorization with temporal smoothness regularization.

\end{abstract}

\section{Introduction}
\label{sec:intro}
Analysis of time-evolving data and revealing the underlying evolving patterns is crucial to understand the functioning of complex systems such as the brain, society, etc. Temporal data can naturally be represented as a higher-order tensor with one of the modes corresponding to time, e.g., neuroimaging signals represented as a \emph{subjects} by \emph{voxels} by \emph{time} tensor \cite{RoShJi19}, time-resolved gut microbiome data as a \emph{subjects} by \emph{microbial features} by \emph{time} tensor \cite{MaSh21}, or  communication data as an \emph{authors} by \emph{terms} by \emph{months} tensor \cite{BaBeBr07a}. 

Tensor factorizations have been successfully used to reveal the underlying patterns of such evolving data. Martino et al. \cite{MaSh21} use the CANDECOMP/PARAFAC (CP) \cite{harshman_foundations_1970, CaCh70} tensor model to analyze gut microbiome data collected over time. The CP model, which can reveal underlying patterns uniquely facilitating interpretation, captures the microbial changes differentiating between infants according to the birth mode in that study. Previously, CP was also used to analyze bibliometric data (i.e., \emph{authors} by \emph{conferences} by \emph{years} tensor), and predict the publications in future years by combining temporal components with time series analysis \cite{dunlavy_temporal_2011}. While these methods preserve the multi-way structure of temporal data and use the time information in subsequent tasks, the said models are not time-aware, i.e., reordering time slices of a tensor results in a reordered version of the same model. To address this issue, recent studies have introduced time-aware tensor factorizations by incorporating temporal regularizers in the time mode \cite{YuRaDh16, TaKaUe17, AhJa22}. Yet, time-aware tensor methods are still limited in terms of revealing evolving patterns. They may reveal static patterns (e.g., words of a topic), and the strength (popularity) of the pattern may change by a scalar encoded in the time component but evolving patterns (e.g., changing groups of terms) cannot be captured.



As an alternative, coupled matrix factorization (CMF)-based approaches, with more modelling flexibility than the CP model, have been effectively used to analyze time-evolving data revealing evolving patterns \cite{YuAgWa17, appel_temporally_2018}. For instance, Temporal Matrix Factorization (TMF) \cite{YuAgWa17} uncovers latent factors that are functions of time. Another method called Chimera \cite{appel_temporally_2018} receives as input two evolving networks and jointly factorizes them assuming smooth changes over time. These methods are primarily used for prediction tasks, and to the best of our knowledge, their uniqueness properties have not been studied. Uniqueness is crucial when the goal is to understand how a complex system evolves in time. 
Streaming/online methods, e.g., \cite{KaLi21,pasricha2018identifying}, can also be used to capture evolving patterns. However, such methods rely on updating patterns in every mode, e.g., both \emph{subjects} and \emph{voxels} modes - which is different than the focus of this study.



In this paper, we introduce the temporal PARAFAC2 (tPARAFAC2) model that combines the advantage of uniqueness with modeling flexibility allowing for evolving patterns. The PARAFAC2 \cite{harshman_parafac2_1972} model allows the patterns in one mode (e.g., voxels, words) to change across slices (e.g., time), and its promise in terms of revealing evolving patterns has recently been studied \cite{RoShJi19}. In order to make PARAFAC2 model time-aware, here, we incorporate a temporal smoothness regularization, assuming smoothly changing patterns over time. Our contributions are as follows:
\begin{itemize}
\vspace{-1mm}
  \item We introduce the tPARAFAC2 model as a time-aware tensor factorization method that can reveal evolving patterns uniquely, and use an Alternating Optimization (AO) - Alternating Direction Method of Multipliers (ADMM) based algorithm to fit the model,
  \vspace{-3mm}
    \item Through extensive experiments on synthetic data, we show that tPARAFAC2 can reveal the underlying evolving patterns accurately, performing better than PARAFAC2 and CMF with temporal smoothness, 
    \vspace{-3mm}
    \item We demonstrate the lack of uniqueness in CMF with temporal smoothness regularization, the advantages of the tPARAFAC2 model as well as its limitations.
\end{itemize}

\section{The proposed method: tPARAFAC2}
\label{sec:proposed_method}
\subsection{CANDECOMP/PARAFAC and PARAFAC2}
\label{ssec:parafac&parafac2}
The CANDECOMP/PARAFAC (CP) model \cite{harshman_foundations_1970,CaCh70} represents higher-order tensors as a sum of rank-one tensors. In the third-order case, an {\small$R$}-component CP model of tensor {\small$\T{X} \in {{\mathbb{R}}^{I \times J \times K}}$} is formulated as
{\small \begin{equation} \label{eq:PARAFAC}
   \T{X} \approx \sum_{r=1}^{R} \MC{A}{r} \circ \MC{B}{r} \circ \MC{C}{r} =  \llbracket \M{A}, \M{B}, \M{C} \rrbracket \, ,
\end{equation}}where {\small$\{\MC{A}{r}\}_{r=1}^{R}$}, {\small$\{\MC{B}{r}\}_{r=1}^{R}$}, {\small$\{\MC{C}{r}\}_{r=1}^{R}$} denote the columns of {\small$\M{A}\in \Real^{I \times R}$}, {\small$\M{B}\in \Real^{J \times R}$} and {\small$\M{C}\in \Real^{K \times R}$}, respectively. Equivalently, the model can be formulated in terms of each frontal slice of {\small$ \bm{\mathscr{X}} $} as follows:
{\small \begin{equation} \label{PARAFAC-slicewise}
   \M{X}_{k} \approx \M{A}\M{D}_{k}\M{B}\Tra \; \forall k \in \{ 1, 2, ... ,K\}
\end{equation}}Here {\small$\M{D}_k \in \mathbb{R}^{R \times R}$} is a diagonal matrix with the {\small$k$}th row of {\small$\M{C}$} on the diagonal. The CP model is unique, under mild conditions, up to permutation and scaling ambiguities \cite{Kr77, KoBa09}.
The PARAFAC2 \cite{harshman_parafac2_1972} model is more flexible compared to CP, as it allows captured patterns in one mode to change across slices. Here, each frontal slice is modeled as
{\small \begin{equation} \label{eq:PARAFAC2-slicewise}
   \M{X}_{k} \approx \M{A}\M{D}_{k}\M{B}_{k}\Tra \; \forall k \in \{1, 2, ... ,K\} \, ,
\end{equation}}where {\small$\M{D}_k$} is again the {\small$R \times R$} diagonal matrix with the {\small$k$}th row of {\small$\M{C}$} on the diagonal and {\small$\{\M{B}_k\}_{k=1}^{K}$} are slice-specific factor matrices. Additionally, the model constrains these matrices to have the same cross product, i.e., {\small$\{\M{B}_k\}_{k=1}^{K}\in\mathscr{P}$} with
{\small\begin{equation} \label{eq:PARAFAC2-constraint}
\mathscr{P} = \left\{ \{\M{B}_k\}_{k=1}^{K} \!\mid\! \M{B}\Tra_{k_{1}}\M{B}_{k_{1}} \!=\! \M{B}\Tra_{k_{2}}\M{B}_{k_{2}} \; \forall k_{1},\! k_{2} \in \{1, 2, ... ,K\}  \right\}.
\end{equation}}We refer to this constraint as the PARAFAC2 constraint. PARAFAC2 is unique up to permutation and scaling under certain conditions, e.g., there are enough slices, nonzero entries in {\small$\M{D}_k$} (see \cite{kiers_parafac2part_1999} for a detailed discussion on uniqueness).

\subsection{tPARAFAC2: Adding temporal regularization}
\label{ssec:adding_temporal_regularization}

When PARAFAC2 is used to analyze time-evolving data with frontal slices {\small$\M{X}_k$} changing in time, {\small$k=1, ..., K$}, the factors {\small$\{\M{B}_k\}_{k=1}^{K}$} can capture the structural dynamics of the patterns across time. Inspired by Appel et al. \cite{appel_temporally_2018}, we incorporate the idea of temporal smoothness in the PARAFAC2 model by constraining consecutive $\M{B}_k$ factors to change smoothly over time, and formulate the optimization problem as follows:
{\small\begin{align}  \label{eq:tPARAFAC2-abstact}
   \min_{\M{A},\{\M{B}_k,\M{D}_k\}_{k=1}^{K}} & \left\{  \sum_{k=1}^{K}\lVert \M{X}_k \!-\! \M{A}\M{D}_k\M{B}\Tra_k\rVert^2_{F} \!+\! g_{\M{A}}(\M{A}) \right. & \notag \\
    &\left. \quad + \lambda_B \sum_{k=2}^K\lVert \M{B}_k \!-\! \M{B}_{k-1} \rVert^2_F \!+\! g_{\M{D}}({{\{\M{D}_k\}}_{k=1}^K}) \right\} \notag \\
    \text{subject to} \; \, \, & \; \{\M{B}_k\}_{k=1}^{K}\in\mathscr{P} ,
\end{align}}where the term {\small$\sum_{k=1}^{K}\lVert \M{X}_k \!-\! \M{A}\M{D}_k\M{B}\Tra_k\rVert^2_{F}$} is the data fitting term and $\norm{.}_{F}$ denotes the Frobenius norm. The penalty terms {\small$g_{\M{A}}$} and {\small$g_{\M{D}}$} denote regularization penalties of the respective factors. The term {\small$\sum_{k=2}^K\lVert \M{B}_k \!-\!\M{B}_{k-1} \rVert^2_F$} corresponds to the ``temporal smoothness" penalty with the penalty parameter {\small$\lambda_{\M{B}}$}.

The optimization problem of \eqref{eq:tPARAFAC2-abstact} is the objective function for our proposed method and yields a PARAFAC2-based decomposition with temporal smoothness. The temporal smoothness constraint is a norm-dependent penalty on the {\small$\M{B}_k$} factors; in order to avoid vanishing factors, we also need to apply regularization to the norm of all remaining factors matrices, hence the existence of {\small$g_{\M{A}}$} and {\small$g_{\M{D}}$} \cite{roald_ao-admm_2022}.
Since the objective function now incorporates time-specific constraints, we name the proposed approach tPARAFAC2.

\subsection{Optimization}
\label{ssec:optimization}

\begin{figure}[t]
\begin{algorithm}[H]
\caption{tPARAFAC2 AO-ADMM}\label{admm-full}
{\small \textbf{Output:} tPARAFAC2 factors $\M{A},\{\M{B}_k,\M{D}_k\}_{k=1}^{K}$.}
\begin{algorithmic}[1]
\State {\small \textit{Initialize} $\M{A}, \{\M{B}_k,\!\M{Y}_{\M{B}_k},\M{Z}_{\M{B}_k},\!\M{\mu}_{\M{Z}_{\M{B}_k}}\!,\!\M{\mu}_{\M{\Delta}_{\M{B}_k}}\!,\!\M{D}_k,\M{Z}_{\M{D}_k},\!\M{\mu}_{\M{D}_k}\}_{k=1}^{K} $}
\State {\small  \textbf{while} \textit{stopping conditions are not met}}
\State {\small \hspace{0.25cm} $ \{{{\M{B}_k},\M{Y}_{\M{B}_k}},\M{Z}_{\M{B}_k},\M{\mu}_{\M{Z}_{\M{B}_k}},\M{\mu}_{\M{\Delta}_{\M{B}_k}}\}_{k=1}^K \gets \text{Algorithm \ref{admm-B}}$}
\State {\small  \hspace{0.25cm} $\{{{\M{D}_k},\M{Z}_{\M{D}_k}},\M{\mu}_{\M{D}_k}\}_{k=1}^K \gets \text{\cite[Algorithm 7]{roald_ao-admm_2022}}$}
\State {\small \hspace{0.25cm} $ \M{A} \gets \argmin_{{\M{A}}} \left\{ \mathcal{L}  \right\}$ \algorithmiccomment{\eqref{eq:A-closed-form}}}
\State {\small  \textbf{end} }
\end{algorithmic}
\end{algorithm}
\vspace{-0.75cm}
\end{figure}

We solve the problem in \eqref{eq:tPARAFAC2-abstact} using the AO-ADMM framework, first introduced for matrix and tensor decompositions in \cite{huang_flexible_2016} and extended for PARAFAC2 in \cite{roald_ao-admm_2022}. Contrary to the standard Alternating Least Squares - based algorithm for PARAFAC2 \cite{kiers_parafac2part_1999}, AO-ADMM makes it possible to impose proximable constraints on any factor, including those subject to the PARAFAC2 constraint \eqref{eq:PARAFAC2-constraint}. We provide an iterative scheme that solves the relevant subproblems of estimating the {\small$\M{B}_k$} factors in tPARAFAC2 and integrate it into AO-ADMM.

Algorithm \ref{admm-full} summarizes our approach. We adopt the same notation as \cite{roald_ao-admm_2022}. At each iteration, we alternatingly solve a convex subproblem for each factor matrix, holding all others fixed.
We use the ridge penalty as the norm-based penalty for the factors of the non-evolving modes. To overcome the sign ambiguity present in PARAFAC2 (and inherently to tPARAFAC2), we also require {\small$\{\M{D}_k\}_{k=1}^K$} to be non-negative. This means that in \eqref{eq:tPARAFAC2-abstact}, we have
\vspace{-0.1cm}{\small \begin{align}
g_{\M{A}}(\M{A})&=\lambda_{\M{A}}\lVert \M{A} \rVert^{2}_F \label{eq:ridge-A} \, ,\\
g_{\M{D}}({{\{\M{D}_k\}}_{k=1}^K})&= \sum_{k=1}^K \left( \lambda_{\M{D}}\lVert \M{D}_k \rVert^{2}_F + \iota_{\mathbb{R}_{+}}(\M{D}_k) \right), \label{eq:ridge-D}
\end{align}}
\vspace{-0.25cm}

\noindent where {\small$\iota_{\mathbb{R}_{+}}$} is the indicator function for the non-negative orthant. For all the non-ridge penalties of \eqref{eq:tPARAFAC2-abstact}, we introduce auxiliary variables {\small $\{\M{Z}_{\M{B}_k},\M{Y}_{\M{B}_k}, \M{Z}_{\M{D}_k}\}_{k=1}^K$}, step-sizes {\small $\{\rho_{\M{B}_k},\rho_{\M{D}_k}\}_{k=1}^K$} and the corresponding feasibility penalties that involve dual variables {\small$\{\M{\mu}_{\bm{Z}_{\M{B}_k}},\M{\mu}_{\bm{\M{\Delta}}_{\M{B}_k}}, \M{\mu}_{\M{Z}_{\M{D}_k}}\}_{k=1}^{K}$}. This yields the augmented Lagrangian
\vspace{-0.1cm}{\small \begin{align} 
    &\mathcal{L} =\sum_{k=1}^{K}\lVert \M{X}_k - \M{A}\M{D}_k\M{B}_k\Tra\rVert^2_{F} + \lambda_{\M{A}}\lVert \M{A} \rVert^{2}_F + \lambda_{\M{D}} \sum_{k=1}^K \lVert \M{D}_k \rVert^{2}_F \notag  \\
    &+ \lambda_B \sum_{k=2}^K\lVert \M{Z}_{\M{B}_k}-\M{Z}_{\M{B}_{k-1}} \rVert^2_F + \sum_{k=1}^K \frac{\rho_{\M{B}_k}}{2}\lVert \M{B}_k -  \M{Z}_{\M{B}_k} + \M{\mu}_{\bm{Z}_{\bm{B}_k}} \rVert^2_{F} \notag \\
    &+ \iota_{\mathscr{P}}(\{\M{Y}_{\M{B}_k}\}_{k=1}^K) + \sum_{k=1}^K \frac{\rho_{\M{B}_k}}{2}\lVert \M{B}_k  - \M{Y}_{\M{B}_k} + \M{\mu}_{\M{\Delta}_{\M{B}_k}} \rVert^2_{F} \notag \\
    &+ \sum_{k=1}^K\iota_{\mathbb{R}_{+}}(\M{Z}_{\M{D}_{k}})  + \sum_{k=1}^K \frac{\rho_{\M{D}_{k}}}{2}\lVert  \M{D}_k  - \M{Z}_{\M{D}_{k}}  + \M{\mu}_{\M{D}_{k}} \rVert^2_{F} ,
    \label{eq:AOADMM-lagrangian}
\end{align}}
\vspace{-0.15cm}

\noindent where {\small$\iota_{\mathscr{P}}$} is the indicator function for set {\small$\mathscr{P}$}. 
The terms in the second line implement the introduced temporal smoothness penalty, while the terms in the third and fourth lines originate from the PARAFAC2 and non-negativity constraints respectively. Minimizing {\small$\mathcal{L}$} w.r.t. {\small$\M{A}$} has a closed form solution:
\vspace{-0.1cm}{\small \begin{equation}
     \M{A}^{*} \!\gets\! \bigl( \sum_{k=1}^K\M{X}_k \M{B}_k \M{D}_k \bigr) \!\bigl( \sum_{k=1}^K (\M{D}_k \M{B}\Tra_k \M{B}_k \M{D}_k) \!+\! \lambda_{\M{A}}\M{I}\bigr)^{\dagger} \label{eq:A-closed-form}
\end{equation}}
\vspace{-0.15cm}

\noindent where {\small$\dagger$} denotes the pseudoinverse. The variables {\small $\M{D}_k, \M{Z}_{\M{D}_k}$}, and {\small $\M{\mu}_{\M{D}_k}$} are updated as in \cite[Algorithm 7]{roald_ao-admm_2022}. For the time-evolving mode, two sets of auxiliary and dual variables need to be updated along with the factor matrices. Algorithm \ref{admm-B} outlines the update procedure. In this case, solving for each {\small$\M{B}_k$} reduces to a problem that admits closed-form solution:
{\small \begin{equation}
\M{B}_{k}^{*} \!\gets\! \bigl( \M{X}_k\Tra \M{A} \M{D}_k + \frac{\rho_{\M{B}{k}}}{2} \M{M} \bigr) \bigl( \M{D}_k \M{A}\Tra \M{A} \M{D}_k + \frac{\M{\rho}_{\M{B}_k}}{2} \M{I}  \bigr)^{\dagger} \label{eq:Bk-closed-form}
\end{equation}}
\vspace{-0.3cm}

\noindent where {\small $\M{M} = \M{Z}_{\M{B}_k} - \M{\mu}_{\M{B}_k} + \M{Y}_{\M{B}_k} - \M{\mu}_{\M{\Delta}_{\M{B}k}}$}. For the auxiliary variables {\small $\M{Y}_{\M{B}_k}$} we employ the approximate projection scheme in \cite{roald_ao-admm_2022}. Setting the derivatives of the lagrangian $\mathcal{L}$ with respect to each {\small $\M{Z}_{\M{B}_1},\M{Z}_{\M{B}_2},\dots,\M{Z}_{\M{B}_K}$} equal to zero yields the following system of equations:
\vspace{-0.05cm} {\small \begin{flalign}
&\bigl( 2\lambda_{\M{B}} + \rho_{\M{B}_1}\bigr) \M{Z}_{\M{B}_1} -2\lambda_{\M{B}}\M{Z}_{\M{B}_2} = \rho_{\M{B}_1} \bigr( \M{B}_1 + \M{\mu}_{\M{B}_1} \bigl) & \label{eq:zb1-derivative}\\
&\bigl(4\lambda_{\M{B}}+\rho_{\M{B}_k}\bigr)\M{Z}_{\M{B}_k}-2\lambda_{\M{B}} \bigl(\M{Z}_{\M{B}_{k-1}}+\M{Z}_{\M{B}_{k+1}} \bigr) \notag \\
&\qquad \qquad=\rho_{\M{B}_k} \bigr(\M{B}_k+\M{\mu}_{\M{B}_k}\bigl), \qquad k = 2,\ldots,K-1
\label{eq:zbk-derivative} \\
&\bigl( 2\lambda_{\M{B}} + \rho_{\M{B}_K}\bigr) \M{Z}_{\M{B}_K} -2\lambda_{\M{B}}\M{Z}_{\M{B}_{K-1}} = \rho_{\M{B}_K} \bigr( \M{B}_K + \,{\mu}_{\M{B}_K} \bigl) \label{eq:zbK-derivative}
\end{flalign}}
\vspace{-0.5cm}

\noindent Here, we solve a block tri-diagonal system of equations using Thomas' method. Lastly, the dual variables are updated according to ADMM (Lines 7-10 of Algorithm \ref{admm-B}). 
\begin{figure}[t]
\begin{algorithm}[H]
\caption{ADMM B Updates}\label{admm-B}
{\small \textbf{Output:} ${\{{{\M{B}_k},\M{Y}_{\M{B}_k}},\M{Z}_{\M{B}_k}},\M{\mu}_{\M{Z}_{\M{B}_k}},\M{\mu}_{\M{\Delta}_{\M{B}_k}}\}_{k=1}^K$.}
\begin{algorithmic}[1]
\State {\small \textbf{while} \textit{stopping conditions are not met}}
\State {\small \hspace{0.25cm} \textbf{for} $k \gets 1 \;\textbf{to}\; K$ \textbf{do}}
\State {\small \hspace{0.55cm} $\M{B}_k \gets \argmin_{{\M{B}_k}} \left\{ \mathcal{L}  \right\}$ \algorithmiccomment{\eqref{eq:Bk-closed-form}}}
\State {\small \hspace{0.25cm} \textbf{end} }
{\small \State \hspace{0.25cm} $ {\{\M{Z}_{\M{B}_K}}\}_{k=1}^K \!\gets\! \text{\emph{Tridiagonal solve}}\! \left(\! \lambda_{\M{B}},\!\{\M{B}_k,\rho_{\M{B}_k}\!,\!\M{\mu}_{\bm{Z}_{\bm{B}_k}}\}_{k=1}^K \right) $}
{\small \State \hspace{0.25cm} $ {\{\M{Y}_{\M{B}_K}}\}_{k=1}^K \!\gets\! \text{\emph{Approx. proj. on}} \, \mathscr{P}\! \left( \{\M{B}_k,\rho_{\M{B}_k}\!,\!\M{\mu}_{\M{\Delta}_{\M{B}_k}}\}_{k=1}^K \right) $}
\State {\small \hspace{0.25cm} \textbf{for} $k \gets 1 \;\textbf{to}\; K$ \textbf{do}}
\State {\small \hspace{0.55cm} $\M{\mu}_{\M{Z}_{\bm{B}_k}} \gets \M{B}_k - \M{Z}_{\M{B}_k} + \M{\mu}_{\M{Z}_{\M{B}_k}} $}
\State {\small \hspace{0.55cm} $\M{\mu}_{\M{\Delta}_{\M{B}_k}} \gets \M{B}_k - \M{Y}_{\M{B}_k} + \M{\mu}_{\M{\Delta}_{\M{B}_k}} $}
\State {\small \hspace{0.25cm} \textbf{end}}
\State {\small \textbf{end} }
\end{algorithmic}
\end{algorithm}
\vspace{-0.75cm}
\end{figure}

As stopping conditions, for the outer ADMM loop (Algorithm \ref{admm-full}) either a pre-set number of iterations has to be reached or an absolute or relative tolerance criterion for the total loss in \eqref{eq:tPARAFAC2-abstact} has to be satisfied. At the same time, the solution has to satisfy the constraints up to a specified tolerance (i.e. feasibility gap).
Inner ADMM loops (e.g. Algorithm \ref{admm-B}) exit either based on maximum iterations or when the absolute or relative distance of the variable to its corresponding ADMM auxiliary variable is smaller than some pre-set tolerance. 

\section{Experimental evaluation}
\label{sec:experimental_evaluation}

\subsection{Synthetic data generation}
\label{ssec:synthetic_data_generation}

In order to evaluate the effectiveness of our proposed method we use a topic modeling-inspired synthetic data generation scheme. Each synthetic dataset can be interpreted as an \emph{authors} $\times$ \emph{words} $\times$ \emph{time} tensor and is composed of a set of artificial concepts evolving over time. In the following experiments, we attempt to track precisely this evolution. In this setting, a concept consists of the following parts: 
\begin{itemize}
    \vspace{-0.2cm}
    \item A set of authors relevant to the concept, which is simulated by a column of {\small$\M{A}$} with values from {\small$\mathcal{N}(0.5,0.5)$} for the relevant authors. Negative values are set to zero.
    \vspace{-0.65cm}
    \item The set of words for each topic and their individual importance at every time slice. We simulate the evolution of concepts with changes across these sets. These are the columns of {\small$\M{B}_k$} factors, where the relevant words at {\small$t=k$} have values from {\small$\mathcal{N}(0.5,0.5)$} and the rest is zero.
    \vspace{-0.25cm}
    \item The strength (i.e., popularity) of a concept at each time point. For each concept, we generate  a sequence of {\small$K$} scalars (columns of {\small$\M{C}$}) and changes are simulated by variations in this sequence. A concept may have \textbf{constant}, \textbf{increasing}, \textbf{decreasing}, or \textbf{peridocially changing} strength (see Fig~\ref{fig:changes}).
    \vspace{-0.25cm}
\end{itemize}

No concepts are evolving in the same way. Influenced by the work of Lu et al. \cite{8496795}, we implemented four kinds of concept drift. Fig.~\ref{fig:drifts} provides visual examples. \textbf{Sudden drift} occurs when a concept changes abruptly. \textbf{Gradual drift} covers the case of fluctuations of the structure of a concept until convergence at a final state. Given a user-specified index $t_0$, we simulate such drift by randomly (Bernoulli distribution with {\small$p=0.5$}) changing between two sets of words {\small $\forall t=\{1,...,t_{0}-1\}$} and then using only the latest word set {\small$\forall t=\{t_{0},...,K\}$}. \textbf{Reoccurring drift} explains the evolution of a concept that alternates between two states periodically. We cycle between two sets of active words according to a user-defined half-period {\small $t_p$}. {\textbf{Incremental drift} refers to the phenomenon in which the concept exhibits a slow transition over time. Such drift is simulated by introducing new words according to a user-defined probability and their individual importance is scaled according to a sigmoid function with user-defined steepness.
\begin{figure}[tb]
    \centering
    \begin{subfigure}[b]{1.0\linewidth}
    \centerline{\includegraphics[width=8.5cm]{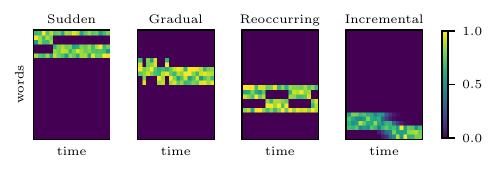}}
      \vspace{-3mm}
      \caption{Four types of concept drift.}
    \end{subfigure}
    \begin{subfigure}[b]{1.0\linewidth}
    \centerline{\includegraphics[width=8.5cm]{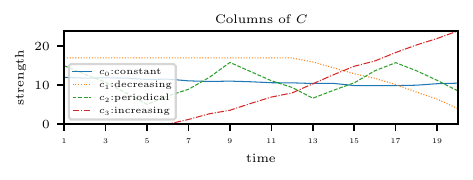}}
      \vspace{-3mm}
      \caption{Four types of strength change.}
      \label{fig:changes}
    \end{subfigure}
    \vspace{-6.5mm}
    \caption{Illustration of (a) concept evolution considered in $\M{B}_k$ matrices, (b) change of strength encoded in $\M{C}$.}
    \label{fig:drifts}
    \vspace{-0.475cm}
\end{figure}

\subsection{Synthetic data experiments}
\label{ssec:synthetic_data_experiments}
In order to evaluate the effectiveness of tPARAFAC2, we have created three groups of synthetic datasets with specific properties. Each dataset is of size {\small$150\times100\times20$} and has three concepts (i.e., {\small$R=3$}). After forming each synthetic tensor, we add noise {\small$\Theta\sim \mathcal{N}(0,1)$} with noise level {\small$\eta$}:
\vspace{-0.1cm} {\small\begin{equation}
    \mathscr{\bm{X}}_{noisy} = \mathscr{\bm{X}} + \eta \lVert \mathscr{\bm{X}}\rVert \frac{\Theta}{\lVert \Theta \rVert}
\end{equation}} 
\vspace{-0.275cm}

\noindent We assess the performance of  tPARAFAC2 in terms of accuracy, i.e., how well the model recovers the ground truth factors used to construct the data. For accuracy, we use the Factor Match Score (FMS) defined as:
\vspace{-0.1cm} {\small\begin{equation}
    \text{FMS} = \sum_{i=1}^R\frac{|\MhatC{A}{i}\Tra\MC{A}{i}|}{\lVert \MhatC{A}{i} \rVert \lVert \MC{A}{i} \rVert} \frac{|\MhatC{B}{i}\Tra\MC{B}{i}|}{\lVert \MhatC{B}{i} \rVert \lVert \MC{B}{i} \rVert}  \frac{|\MhatC{C}{i}\Tra\MC{C}{i}|}{\lVert \MhatC{C}{i} \rVert \lVert \MC{C}{i} \rVert},
\end{equation}}
\vspace{-0.275cm}

\noindent where {\small$\MC{A}{i}$}, {\small$\MC{B}{i}$} and {\small$\MC{C}{i}$} denote the ground truth components and {\small$\MhatC{A}{i}$}, {\small$\MhatC{B}{i}$} and {\small$\MhatC{C}{i}$} their estimates. We compare tPARAFAC2 with PARAFAC2 as well as with a time-aware CMF model that has no PARAFAC2 constraint and is formulated as follows: \vspace{-1mm}
{\small\begin{equation} \label{eq:tCMF-abstact}
    \argmin_{\M{A},\{\M{B}_k\}_{k=1}^{K}} \left\{  \sum_{k=1}^{K}\lVert \M{X}_k \!-\! \M{A}\M{B}_k\Tra\rVert^2_{F} + \lambda_B \sum_{k=2}^K\lVert \M{B}_k \!-\! \M{B}_{k-1} \rVert^2_F \right\}
\end{equation}}
\vspace{-0.3cm}

\noindent We refer to \eqref{eq:tCMF-abstact} as tCMF and when computing FMS, we omit the last term of each summand. We use AO-ADMM for fitting all models, with the number of components set to {\small$R=3$}. Implementation\footnote{https://github.com/cchatzis/tPARAFAC2} was done in Python using TensorLy \cite{tensorly} and MatCoupLy \cite{ROALD2023101292}.

For each dataset, we initialized all methods with the same 20 random initializations from {\small$\mathcal{U}(0,1)$}, and the run with the lowest total loss is considered for assessing the performance. Runs that reached the maximum number of iterations or converged to degenerate solutions \cite{ZiKi02} (i.e., where two components are highly correlated in all modes but point to different directions) were discarded.

\begin{figure}[tb]
    \begin{minipage}[b]{1.0\linewidth}
      \centering
      \centerline{\includegraphics[width=1.0\textwidth]{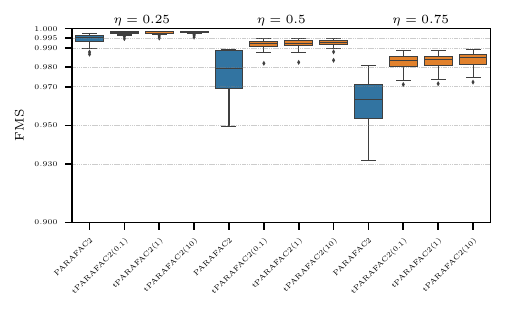}}
      \vspace{-4mm}
    \caption{Easy case. Accuracy in terms of FMS of PARAFAC2 and tPARAFAC2 at different noise levels. The number next to the method name denotes $\lambda_{\M{B}}$. Each boxplot shows the distribution of FMS values for the best-performing runs of each method for 20 datasets.}
    \label{fig:easycase}
    \end{minipage}
     \vspace{-0.95cm}
\end{figure}
\begin{figure*}[htp]
    \begin{minipage}[b]{1.0\linewidth}
      \centering
      \centerline{\includegraphics[width=1.0\textwidth]{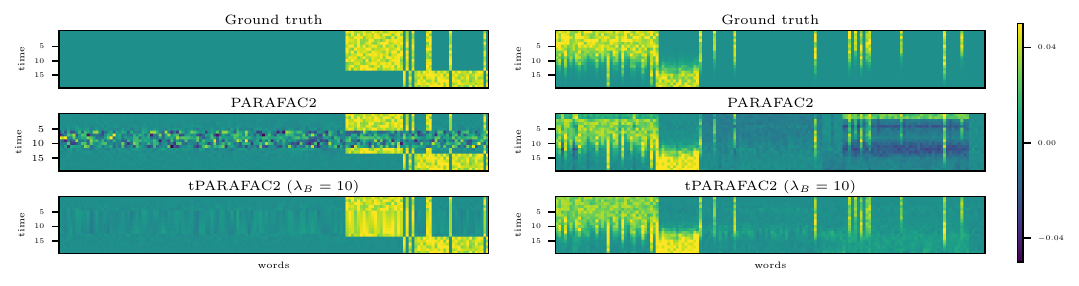}}
      \vspace{-4mm}
      \caption{Example ground truth and recovered columns of {\small$\M{B}_k$} from concepts from the \textit{near zero} (left column) and \textit{overlapping concepts - 20\% overlap} (right column) cases. Each heatmap shows the evolution of a concept over time. }
    \label{fig:nearzero_example}
    \end{minipage}
    \vspace{-0.6cm}
\end{figure*}

\begin{figure}[htb]
    \begin{minipage}[b]{1.0\linewidth}
      \centering
      \centerline{\includegraphics[width=1.0\textwidth]{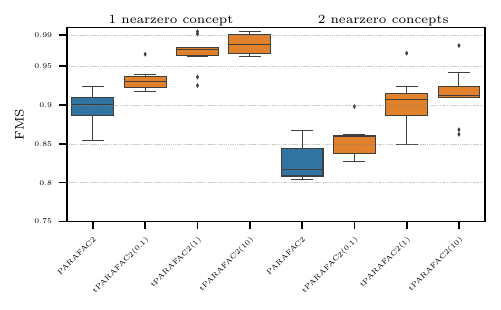}}
      \vspace{-4mm}
      \caption{Almost Zero case. The boxplots show the distribution of FMS with the ground truth factors for the best-performing runs of each method for the 20 datasets of the case.}
    \label{fig:nearzero}
    \end{minipage}
    \vspace{-0.99cm}
\end{figure}
\vspace{-0.04cm}
\textbf{Easy case}: The first group consists of 20 datasets with three concepts that evolve according to \textit{sudden}, \textit{gradual}, or \textit{reoccurring drift}. The indices {\small$t_0$}  and the period {\small$t_p$} are chosen randomly from {\small$\mathcal{U}(2,18)$}. The concept strength is well above zero and chosen to be either constant, periodical, or decreasing. Note that in this case, the PARAFAC2 constraint holds for the ground truth. Fig.~\ref{fig:easycase} shows that, as noise increases, we notice a decrease in the recovery performance of the ground truth components. However, tPARAFAC2 is more robust to noise than PARAFAC2 and achieves higher FMS values.

\textbf{Almost zero strength}: In this case, we evaluate the performance of tPARAFAC2 when the strength of a concept becomes very small. For example, that could be the case when a new concept may be slowly forming. In PARAFAC2, small values on the diagonal of the {\small$\M{D}_k$} factors hinder the quality recovery of the respective columns of {\small$\M{B}_k$} \cite{roald_ao-admm_2022}. We created 20 datasets with one and two of the concepts present respectively (40 in total) having strength as low as {\small$0.01$} for 4-6 slices. The drift type is chosen as in the \textit{easy case} and the PARAFAC2 constraint is satisfied. The results are shown in Fig.~\ref{fig:nearzero}. PARAFAC2 is unable to correctly estimate the columns of the {\small$\M{B}_k$} at the time slices with concept strength close to zero. Instead, the factors accumulate noise, which results in reduced accuracy. tPARAFAC2 is able to overcome this difficulty and provide a more accurate and interpretable result. The left column of Fig.~\ref{fig:nearzero_example} shows an example of how the smoothness constraint assists in capturing the evolving patterns more accurately.

\begin{figure*}[htp]
    \begin{minipage}[b]{1.0\linewidth}
      \centering
      \centerline{\includegraphics[width=1.0\textwidth]{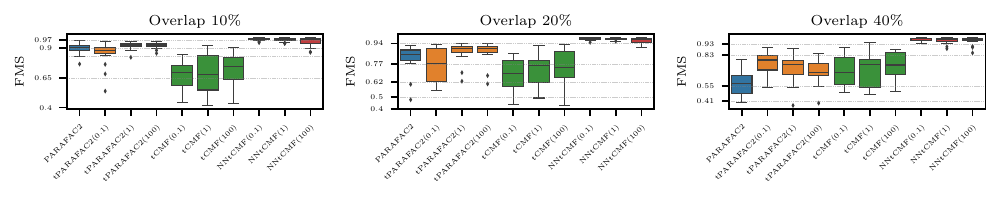}}
      \vspace{-4mm}
      \caption{FMS of all methods in the \emph{Overlapping concepts} case, with noise level {\small$\eta=0.5$} and different ({\small$\%$}) of change in overlap. Each boxplot shows the FMS values for the best-performing runs (lowest total loss) of each method for 20 datasets.}
    \label{fig:overlapping}
    \end{minipage}
    \vspace{-0.65cm}
\end{figure*}

\textbf{Overlapping concepts}: We now investigate the case where the ground truth does not follow the PARAFAC2 constraint. In this, all concepts evolve according to incremental drift. Constraining the inner product of the temporal factors to be constant requires (a) the norm of each column of {\small$\M{B}_k$} as well as (b) the angle between different columns to stay unchanged across the time-slices. Here, (a) is violated due to the scaling imposed by the incremental drift. To violate (b), we force all concepts to share a percentage of words at {\small$t=1$} (i.e., some overlap of words) while making sure they eventually become orthogonal (no overlap). We create three sets of 20 datasets, with {\small$10\%$}, {\small$20\%$} and {\small$40\%$} of the words initially shared, in order to assess the performance at different levels of change in overlap.  Fig.~\ref{fig:overlapping} shows the performance of the methods with tPARAFAC2 achieving higher FMS compared to PARAFAC2 and revealing the underlying patterns despite the violation of the PARAFAC2 constraint. Many of the factors of PARAFAC2 solutions contain more than one concept (see right column of Fig.~\ref{fig:nearzero_example}). The irrelevant concept is an artifact of the PARAFAC2 constraint. The smoothness constraint assists in alleviating that, while recovery of {\small$\M{A}$} and {\small$\M{D}_k$} also improves. As the change in overlap increases, it gets increasingly difficult to satisfy the PARAFAC2 constraint, hence the decrease in the performance of both tPARAFAC2 and PARAFAC2. Even though it possesses more structural flexibility, tCMF has low FMS scores in all cases; that is due to the lack of uniqueness of the tCMF solution. Therefore, recovery cannot be improved simply by omitting the PARAFAC2 constraint. Here, we also consider imposing non-negativity constraints in {\small$\M{A}$} in \eqref{eq:tCMF-abstact}, and refer to the model as NNtCMF in Fig.~\ref{fig:overlapping}. We can see that in the present case, adding a non-negativity constraint yields unique solutions and much-improved recovery of the factors. However, imposing non-negativity constraints is not always possible in applications, especially when the data is centered, e.g., across the \emph{subjects} mode.


\section{Conclusions}
\label{sec:conclusion}

In this work, we have introduced tPARAFAC2, a PARAFAC2-based decomposition with temporal regularization. Using time-evolving synthetic data, we have demonstrated the effectiveness of tPARAFAC2 in terms of tracking slowly evolving patterns, especially in noisy cases and when patterns are more prone to noise due to low signal strength. Although uniqueness is crucial for interpretability, and the proposed approach achieves that through the PARAFAC2 constraint, our experiments show that when the evolution of the patterns violates the constraint significantly, the recovery performance degrades. While more flexible decompositions such as CMF may be more effective, they fail to reveal unique patterns just with the temporal smoothness regularization and require additional constraints. 


\section{Acknowledgements}
\label{sec:acknowledgements}

We would like to express our gratitude to Marie Roald, Carla Schenker, Lu Li, and Shi Yan for their assistance and fruitful discussions. This work was supported by the Research Council of Norway through project 300489 and benefited from the Experimental Infrastructure for Exploration of Exascale Computing (eX3) under contract 270053.

\bibliography{Latex_Template/mlsp_template_camera_ready.bib}

\end{document}

%% file: pre-packages.tex
\usepackage{amsmath}

\usepackage{amsfonts}

\usepackage{amssymb}

\usepackage{stmaryrd}

\usepackage[mathscr]{eucal}

\usepackage{amsbsy}

\usepackage{bm}

\usepackage{bbding}

\usepackage{array}

\usepackage{tabularx}

\usepackage{ctable}

\usepackage[neveradjust]{paralist}

\usepackage{fancyvrb}

\usepackage{caption}

\usepackage{placeins}

\usepackage{ifpdf}
\ifpdf
\DeclareGraphicsExtensions{.pdf,.png}
\else
\DeclareGraphicsExtensions{.eps}
\fi

\usepackage{algpseudocode}


%% file: pre-declarations.tex
\newcommand{\email}[1]{\href{mailto:#1}{\nolinkurl{#1}}}

\newcommand{\Sec}[1]{\hyperref[sec:#1]{\S\ref*{sec:#1}}} 
\newcommand{\Section}[1]{\hyperref[sec:#1]{Section~\ref*{sec:#1}}} 
\newcommand{\AppFull}[1]{\hyperref[sec:#1]{Appendix~\ref*{sec:#1}}} 
\newcommand{\Eqn}[1]{\hyperref[eq:#1]{(\ref*{eq:#1})}} 
\newcommand{\Fig}[1]{\hyperref[fig:#1]{Figure~\ref*{fig:#1}}} 
\newcommand{\Tab}[1]{\hyperref[tab:#1]{Table~\ref*{tab:#1}}} 
\newcommand{\Thm}[1]{\hyperref[thm:#1]{Theorem~\ref*{thm:#1}}} 
\newcommand{\Cor}[1]{\hyperref[cor:#1]{Corollary~\ref*{cor:#1}}} 
\newcommand{\Alg}[1]{\hyperref[alg:#1]{Algorithm~\ref*{alg:#1}}} 
\newcommand{\Def}[1]{\hyperref[def:#1]{Definition~\ref*{def:#1}}} 

\newcommand{\Real}{{\mathbb R}}



%% file: pre-definitions.tex
\newcommand{\Tra}{^{{\sf T}}} 
\newcommand{\V}[1]{{\bm{\mathbf{\MakeLowercase{#1}}}}} 
\newcommand{\Vhat}[1]{{\bm \hat{\mathbf{\MakeLowercase{#1}}}}} 

\newcommand{\M}[1]{{\bm{\mathbf{\MakeUppercase{#1}}}}} 
\newcommand{\MC}[2]{\V{#1}_{#2}} 
\newcommand{\MhatC}[2]{\Vhat{#1}_{#2}} 

\newcommand{\T}[1]{\boldsymbol{\mathscr{\MakeUppercase{#1}}}} 

\newcommand{\norm}[1]{\left\lVert \, #1 \, \right\rVert}

